\documentclass{svproc}
\usepackage{url}

\usepackage[utf8]{inputenc}
\usepackage[T1]{fontenc}
\usepackage{graphicx}
\usepackage{grffile}
\usepackage{longtable}
\usepackage{wrapfig}
\usepackage{rotating}
\usepackage[normalem]{ulem}
\usepackage{amsmath}
\usepackage{textcomp}
\usepackage{amssymb}
\usepackage{capt-of}
\usepackage{hyperref}
\usepackage{dsfont}
\usepackage{comment}
\usepackage{rotating}
\usepackage{color}
\usepackage{subcaption}
\usepackage{todonotes}
\usepackage{algorithm}
\usepackage[noend]{algpseudocode}

\def\ic!#1!{{\footnotesize [#1]}}
\def\icn!#1!{{\footnotesize #1}}

\def\fakesc{\setlength{\currentsize}{\f@size pt}\bgroup\obeyspaces\fakesci}

\hypersetup{colorlinks=true,linktoc=all}
\DeclareMathAlphabet      {\mathbff}{OT1}{cmr}{bx}{it}

\begin{document}
\mainmatter              
\title{Combinatorics of a Discrete Trajectory Space for Robot Motion Planning}
%
%
\author{Felix Wiebe\inst{1,\star} \and
Shivesh Kumar\inst{1} \and
Daniel Harnack\inst{1} \and
Malte Langosz\inst{1} \and
Hendrik Wöhrle\inst{1,2} \and
Frank Kirchner\inst{1,3}}
\authorrunning{Felix Wiebe et al.} 
%
%
\institute{Deutsches Forschungszentrum fuer Kuenstliche Intelligenz GmbH, Robotics Innovation Center, Bremen, Germany \and
Institute for Communication Technology, University of Applied Sciences and Arts, Dortmund, Germany \and
AG Robotics, University of Bremen, Germany\\ 
\email{\inst{\star} felix.wiebe@dfki.de}}

\maketitle              

\begin{abstract}
Motion planning is a difficult problem in robot control. 
The complexity of the problem is directly related to the dimension of the robot's configuration space. 
While in many theoretical calculations and practical applications the configuration space is modeled as a continuous space, 
we present a discrete robot model based on the fundamental hardware specifications of a robot. 
Using lattice path methods, we provide estimates for the complexity of motion planning
by counting the number of possible trajectories in a discrete robot configuration space. 

\end{abstract}
%
\setcounter{tocdepth}{5}

\hypersetup{linkcolor=black}

\section{Introduction}

Motion planning is one of the most challenging problems in robotics and is NP-hard in general. 
The central object in the motion planning problem is the configuration space (\emph{c-space}) of the robot, 
which is usually modeled as a continuous space and is further equipped with the structure of a differentiable manifold.  
In a continuous space, the number of unique trajectories is not countable as they can differ by arbitrary small displacements. 
There are several papers that describe the complexity of motion planning. 
For example,~\cite{farber2003topological} describes the notion of topological complexity 
which measures discontinuity of the motion planning process in a given \emph{c-space}. 
In~\cite{erickson2013simple}, it is proven that even a simple two dimensional motion planning problem in $\mathbb{R}^2$ is NP-hard.
Nevertheless, modeling the \emph{c-space} as a continuous and differentiable manifold has numerous advantages 
as it allows the use of various powerful tools from differential geometry and optimization based methods 
to find an optimal path for the robot under various constraints. 

%
However, most of today's measuring devices themselves are actually \emph{digital}, 
which means the results of their measurements have a certain accuracy that is determined by the device itself and all measurements are only provided as discrete values.
On top of that, the measured values are usually processed by some computing device that can only provide approximate real numbers by floating or fixed point numbers.
Consequently, the digital representation of the state space of a robot itself is a discrete space.
Discrete mathematics provides another theoretical perspective and alternative tools to deal with certain problems.
For example, in state lattice planning \cite{pivtoraiko2009differentially}, kinodynamically feasible paths are stored in a graph structure. 
For motion planning there is a trade off between resolution and computation efficiency. 

In this paper, we analyse the complexity of the motion planning problem for a discrete robot model where the state space resolution is not chosen to meet computation requirements but derived from the hardware limitations of the robot itself. 
In a discrete configuration space, the number of trajectories is not only countable but even finite (for a given length).
We discuss the combinatorics of the trajectory space in order to determine how large it can be and 
and how it scales with the parameters describing the robot's structure. 
This theoretical result can be used to assess the accuracy of parametrised approaches that assume a continuous \emph{c-space}.

\section{Discrete Configuration Space Model}
This section explains the discrete robot model for a serial robot.
We assume that in a robotic system most components operate with digital inputs and outputs at fixed frequencies.
This means that they receive or send signals at fixed time steps. 
Therefore, we model time at the fundamental level as a discrete parameter.

\begin{definition}{Time {\normalfont and} Temporal Resolution}\\
The \emph{temporal resolution} is given by the inverse of the lowest frequency of all \emph{operating frequencies} $(f_1, f_2, ...)$ of the robot:
\vspace{-8pt}
\begin{equation}
 \delta t = \frac{1}{\min_j(f_j)}
 \vspace{-8pt}
\end{equation}
Consequently, \emph{time} is an element of a discrete set $t \in \{0, \, \delta t, \,2 \delta t, \, ...\} = T$, 
where the clock starts at $t=0$.
\end{definition}

\begin{definition}{Robot Configuration {\normalfont and} Spatial Resolution}\\
A single joint can be in a position $q \in Q$, where $Q$ is the set of all possible positions for that joint and is usually bounded by the joint limits.

The robot model with $n$ joints has a \emph{configuration} $\pmb{q} = [q_1, q_2, ..., q_{n}] \in \pmb{Q}$, where $q_i \in Q_i$. 
The configuration is an element of the configuration set 
$\pmb{Q} \subseteq Q_1 \times Q_2\times ... \times Q_{n}$ which is a subset of the Cartesian product of the individual joints' configuration spaces.

Two configurations $\pmb{q}^a$ and $\pmb{q}^b$ are distinguishable only if 
at least one joint position differs by more than its \emph{spatial resolution} $\delta q_i$:
\vspace{-4pt}
\begin{equation}
 \exists i \in \{1, 2, ..., n\} \, :\, ||q^b_i - q^a_i|| \geq \delta q_i
 \vspace{-4pt}
 \end{equation}
\end{definition}

This implies that the configuration set is a discrete set. The spatial resolution is introduced by the resolution of the sensors measuring the configuration. 


\begin{definition}{Joint Velocity}\\
The \emph{joint velocity} is the discrete time derivative $\dot{\pmb{q}} = \frac{\pmb{q}_{t+1} - \pmb{q}_t}{\Delta t} \in \pmb{\dot{Q}}$ of the configuration.
\end{definition}

\begin{definition}{Kinematic Robot State}\\
The \emph{kinematic state} of the robot is a tuple of the configuration and the joint velocity: 
$ \pmb{s} = (\pmb{q}, \dot{\pmb{q}}) \in \pmb{S} = \pmb{Q} \times \pmb{\dot{Q}}$.
\end{definition}

For simplicity, we assume that the robot has $n$ actuators which are directly built into the joints. Furthermore, we assume that the actuators are controlled by motors and can be actuated independently.

\begin{definition}{Motor Output}\\
The motors itself receive a digital PWM signal as input and output a force or torque 
$u \in \{u_{min}, \, ...\, , -\delta u, \,0 \,, \,\delta u, \,...\,, u_{max}\} = \mathcal{U}$. 
The force/torque set $\mathcal{U}$ is bounded by the motor limits. 
\end{definition}

Introduced by the digital nature of the input, the force/torque output space is discrete with a resolution $\delta u$. 
The value of $\delta u$ is given by the motor hardware and the way the input signal is modeled.
The force or torque applied to the $i$'th actuator is $u_i \in \mathcal{U}_i$ with a resolution $\delta u_i$. 
The complete force/torque set is
\vspace{-4pt}
\begin{equation}
 \pmb{u} = [u_1, u_2, ..., u_{n}] \in \pmb{\mathcal{U}} = \mathcal{U}_1 \times \mathcal{U}_2 \times ... \times \mathcal{U}_{n}
 \vspace{-4pt}
\end{equation}

With the motor output we now define an action of the robot:

\begin{definition}{Action}\\
We call $\pmb{A} = \pmb{\mathcal{U}} \times T$ the action space. An element $\pmb{a} = (\pmb{u}, \Delta t) \in \pmb{A}$ is an \emph{action} and consists of a force/torque vector 
$\pmb{u} \in \pmb{\mathcal{U}}$ and a time duration $\Delta t \in T$.
\end{definition}

\begin{definition}{Actuation}\\
An \emph{actuation} is an action applied to the robot in a kinematic state $\pmb{s} = (\pmb{q}, \dot{\pmb{q}}) \in \pmb{S}$. 
The outcome of an actuation is a new state of the robot.
In this framework an actuation is a mapping
\begin{equation}
\text{act} : \pmb{S} \times \pmb{A} \rightarrow \pmb{S}
\end{equation}
The outcome of this function is defined by the equations of motion for the robot and reflects its forward dynamics. 
The actuation of a sequence of actions $\hat{\pmb{a}} = [\pmb{a}^0, \pmb{a}^1, ..., \pmb{a}^{m-1}] \in A^m$ applied to the start state $\pmb{s}^0 \in \pmb{S}$ can be defined as a recursive function $Act_m : \pmb{S} \times \pmb{A}^m \rightarrow \pmb{S}$ with
\vspace{-4pt}
\begin{equation}
 Act_m (\pmb{s}^0, \hat{\pmb{a}}) = 
 \begin{cases}
 \begin{alignedat}{2}
  &act(\pmb{s}^0, a^0) \hspace{2.25cm} \quad, m=1
  \\
  &act(Act_{m-1}(\pmb{s}^{0}, \hat{\pmb{a}}), a^{m-1}) \quad, m>1
  \end{alignedat}
 \end{cases}
 \vspace{-4pt}
\end{equation}

\noindent A triple $(\pmb{s}^{a}, \hat{\pmb{a}}, \pmb{s}^{b})$ where $\pmb{s}^b = \text{Act}_m(\pmb{s}^a, \hat{\pmb{a}})$ is called a \emph{transition}.
\end{definition}

We will call the smallest building blocks for constructing more complex motions `atomic actions'. 

\begin{definition}{Time Atomic Action}\\
An action $\pmb{a} = (\pmb{u}, \Delta t)$ is \emph{time atomic} if the action's time duration is equal to the time resolution $\Delta t = \delta t$.
\end{definition}
It can happen that the effect of a time atomic action is not measurable because the change in configuration space is too small. For this reason they are not ideal as atomic actions. Instead, we define atomic actions as:

\begin{definition}{Atomic Action}\\
 A sequence of time atomic actions $\hat{\pmb{a}} = [\pmb{a}^0, \pmb{a}^1, ..., \pmb{a}^{m-1}] \in \pmb{A}^m$ is called an atomic action at $\pmb{s}^0 = (\pmb{q}^0, \dot{\pmb{q}}^0) \in \pmb{S}$ if for $Act_m(\pmb{s}^0, \hat{\pmb{a}}) = \pmb{s}^m = (\pmb{q}^m, \dot{\pmb{q}}^m)$ holds that:
\begin{enumerate}
  \item \begin{enumerate}
         \item $\exists i: ||q^m_i-q^0_i|| \geq \delta q_i$\\
         or
         \item $\text{For } \hat{\pmb{a}}^M = M \cdot [\pmb{a}^0, ..., \pmb{a}^{m-1}]  \text{ and } Act_{M\cdot m}((\pmb{q}^0, \dot{\pmb{q}}^0), \hat{\pmb{a}}^M) = (\pmb{q}^{M\cdot m}, \dot{\pmb{q}}^{M \cdot m}) \\ \text{holds } ||q^{M \cdot m}_i - q^0_i|| < \delta q_i \quad ,\forall M \in \mathbb{N}, i$
        \end{enumerate}
  \item $\forall m'<m, i: ||q^{m'}_i - q^0_i|| < \delta q_i$
 \end{enumerate}
 The condition 1.(a) enforces that the configuration changes in at least one joint while the condition 1.(b) includes the `null action' which holds the current position with zero velocity. The multiplication with the sequence is to be understood as repeating the sequence $M$ times. The last condition makes sure that the atomic action ends as soon as a change in configuration space has been detected.
\end{definition}

\begin{figure}[h]
\centering
\includegraphics[width=0.5\textwidth]{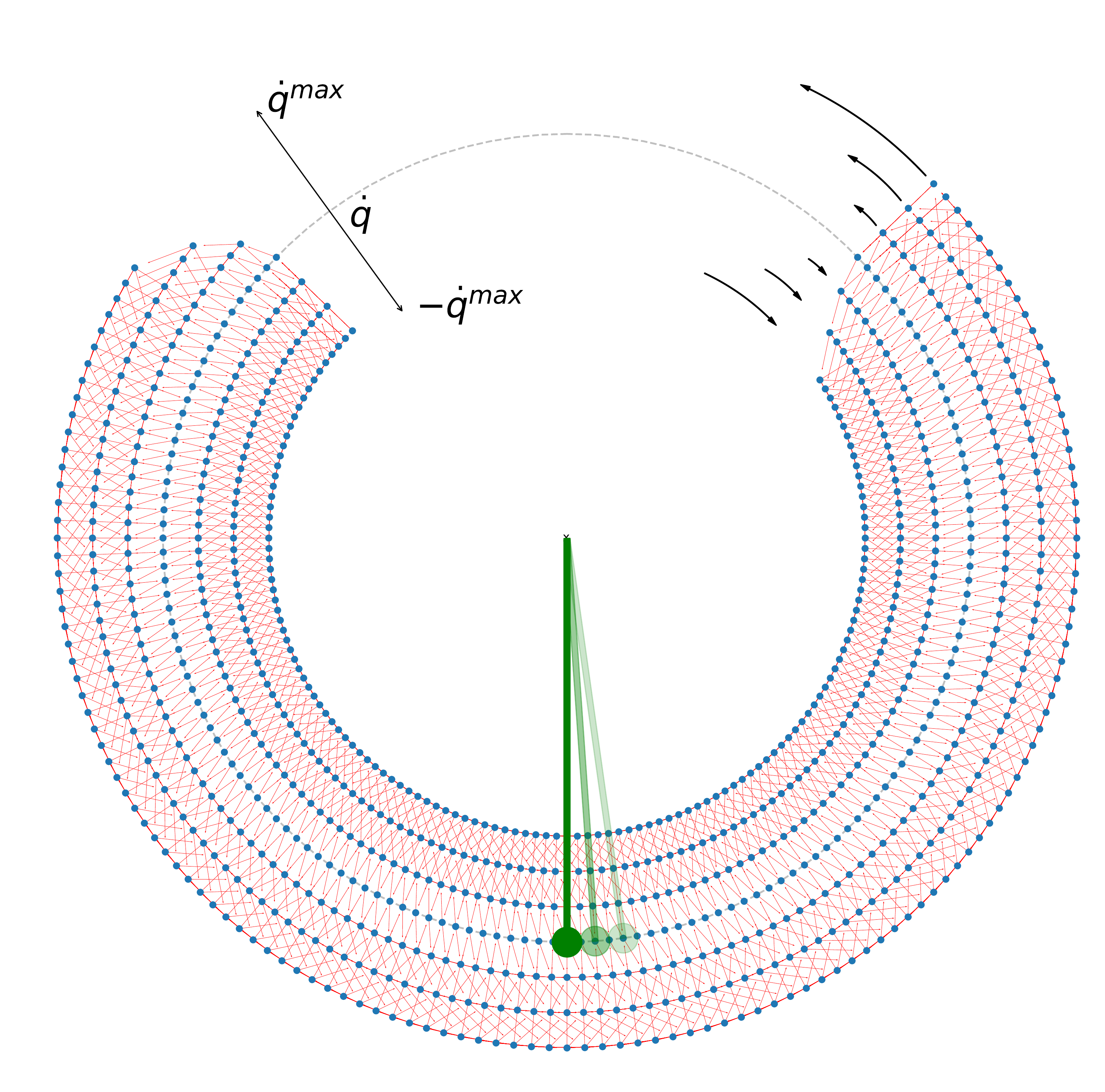}
\includegraphics[width=0.28\textwidth, angle=270,origin=c]{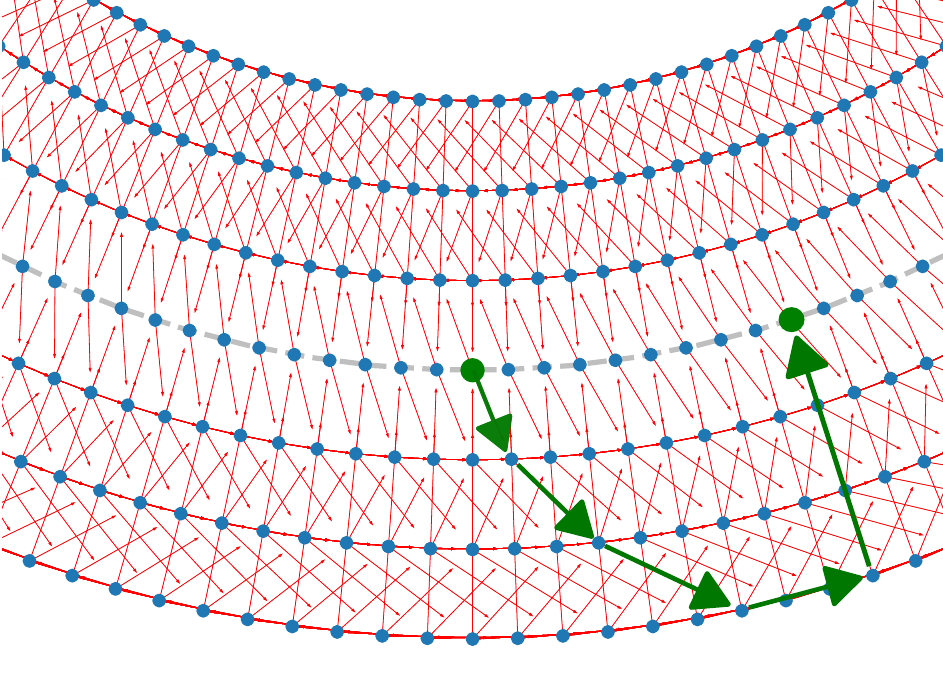}
\vspace{-6pt}
\caption[caption]{Left: Transition map of a single joint. The blue dots are states and the red arrows depict atomic transitions between states. The visualisation uses polar coordinates  where the angle describes the position and the radius scales with the velocity.
Right: Enhanced view of the atomic transition map. The green arrows indicate one possible path from the left green state to the right green state.}
\label{fig:TransitionMapTimeMin}
\vspace{-15pt}
\end{figure}

\begin{example}{Transition map of a single joint}\\
Figure \ref{fig:TransitionMapTimeMin} (left) shows the kinematic state space and time atomic actions for a single joint. The joint has been modeled as a mathematical pendulum with length $l=1\,\text{m}$ and a mass of $m = 1 \, \text{kg}$ at the tip. The joint has the hardware specifications: $Q = [-135^{\circ}, 135^{\circ}]$, $\delta q = 2^{\circ}$, $\dot{Q} = [-180^{\circ}\,\text{s}^{-1}, 180^{\circ}\,\text{s}^{-1}]$, $\delta t = 40 \,\text{ms}$, $\mathcal{U} = \{-50, -25, 0, 25, 50\}[\text{Nm}]$. The outcome of the actuations has been calculated by solving the equations of motion numerically.
\end{example}

Chaining actions and applying them to the robot results in a trajectory:
\begin{definition}{Trajectory}\\
A trajectory is a temporal sequence (an ordered set) of waypoints which are (state, time)-tuples:
\vspace{-4pt}
\begin{equation}
 l^{m} = [(\pmb{s}^0, t^0), (\pmb{s}^1, t^1), ..., (\pmb{s}^{m-1}, t^{m-1})]
 \vspace{-4pt}
\end{equation}
where the time coordinates are ordered: $t^0 < t^1 < ... < t^m$. 
The length of the trajectory is equal to the number of steps $m$ and the space of all feasible trajectories of length $m$ is $L^{m} \subset (\pmb{S} \times T)^{m}$.
\end{definition}

\section{Combinatorics of the Trajectory Space}

Let the number of configurations of joint $i$ be $|Q_i|$ and the number of possible joint velocities be $|\dot{Q}_i|$. The total number of states for the whole system is
\vspace{-4pt}
\begin{equation}
|\pmb{S}| = \prod_{i=1}^{n} |Q_i| |\dot{Q}_i|
\vspace{-4pt}
\end{equation}

In each state the robot has the atomic actions associated to that state available. As an estimate we can assume that the set of atomic actions is the same for all states. If the number of atomic actions of joint $i$ is $|A_i|$, the total number of actions that can be chosen for the robot is
\vspace{-4pt}
\begin{equation}
 |\pmb{A}| = \prod_{i=0}^{n} |A_i|
 \vspace{-4pt}
\end{equation}

The number of unique trajectories is the product of the number of starting states and the available atomic actions to the power of the number of steps $m$:
\vspace{-4pt}
\begin{equation}
 |L^m| =   |\pmb{S}| |\pmb{A}|^{m}
 \vspace{-4pt}
\end{equation}
So the number of possible states, the number of available actions and the number of unique trajectories all scale exponentially with the number of joints. This is often referred to as the curse of dimensionality. The number of trajectories additionally increases exponentially with the trajectory length.
All trajectories can be composed of the same elementary building blocks which are transitions with atomic actions (the $L^1$ space). 

This calculation does not respect joint limitations. Additionally, some actions may have the same outcome as their result can be measured only on the discrete configuration grid. Lastly, not all actions are available at every state as they can produce undesired self-collisions. 



However, this simple analysis gives an upper bound for the number of trajectories and correctly describes the scaling of this quantity with the fundamental parameters $|\pmb{S}|$ and $|\pmb{A}|$. Another interesting question in the context of trajectory planning is: `How many trajectories are possible if the start and end points are fixed?'. For this we use results from the field of lattice paths \cite{ault2019counting}.\\
\indent A lattice path is a path on a discrete grid. A position on the $n$-dimensional grid is described by integers $\mathbb{Z}^n$ and the steps $\mu$ that can be taken on the grid are defined in a so called move set $M \subset \mathbb{Z}^n$. A path is a sequence of positions $(z_0, z_1, ..., z_m) \in \mathbb{Z}^{n \times m}$ where $\forall i > 0: \, z_i - z_{i-1} \in M$. This can be seen as an agent moving on the grid by taking a step from the move set at every time step.\\
\indent A trajectory on the configuration grid can be modeled as a lattice path. Neglecting velocities, the move set $M$ will contain all steps to the nearest neighbors for each joint. 
With the assumption that the joints can be moved independently, diagonal steps are also allowed.
So the move set of robot with $n = 2$ would be
\vspace{-4pt}
\begin{equation}
 M_2 = \{ (0,0), (\pm \delta q_1, 0), (0, \pm \delta q_2), (\pm \delta q_1, \pm \delta q_2)\}
 \label{eq:stepset}
 \vspace{-4pt}
\end{equation}
So there are $|M_2| = 9$ steps in total. Eight of them to the nearest neighbors and one step that does not change the position. For $n$ degrees of freedom (dof) there are $|M_{n}| = 3^{n}$ steps.

\begin{wrapfigure}{r}{0.5\textwidth}
\vspace{-12pt}
\includegraphics[width=0.5\textwidth]{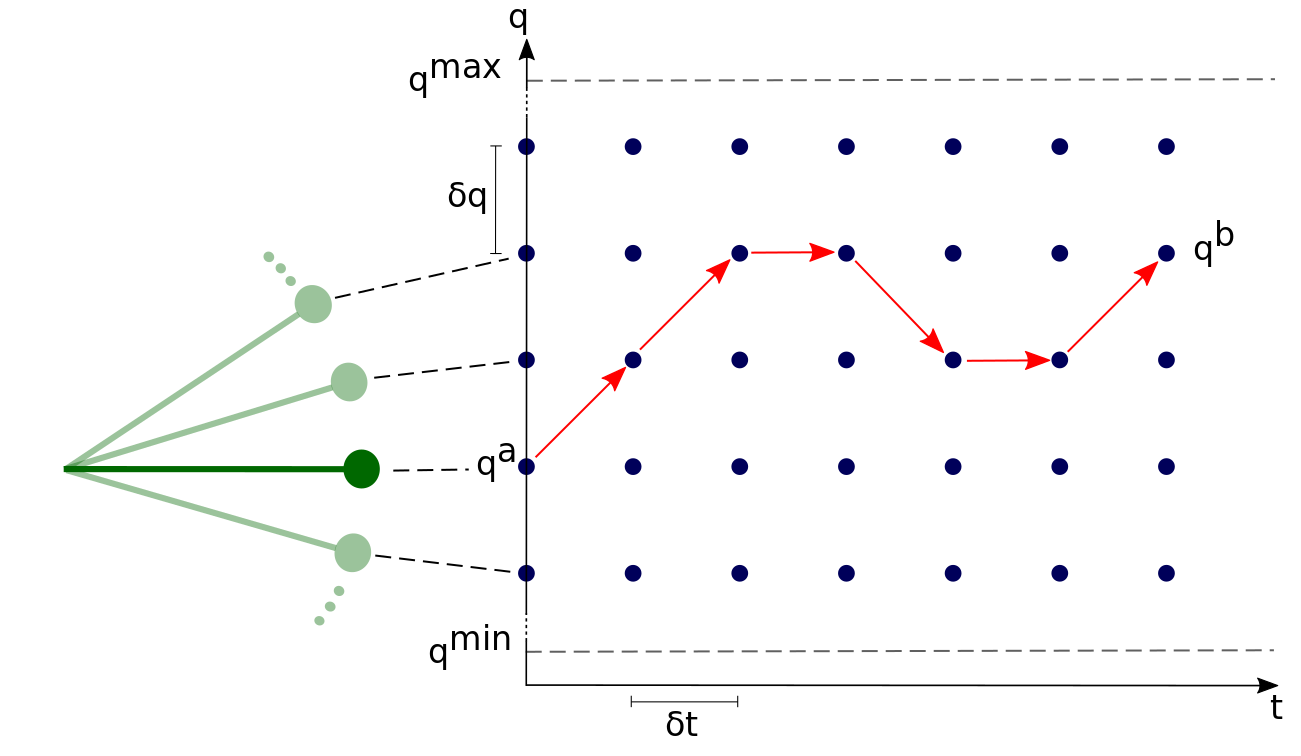}
\caption{The motion of a single joint as a 3-way corridor.}
\label{fig:grid_corridor}
\vspace{-25pt}
\end{wrapfigure}
In order to achieve an analytical representation for the number of paths we consider a version of lattice paths which are called corridors.
A corridor (see also \cite{ault2019counting}) has only one spatial dimension and is described by a tuple $(z, j) \in \mathbb{Z} \times \{ 0, 1, ..., m\}$. $z$ denotes the position and $j$ the step counter. Additionally, a corridor can have a height $h$ which sets a limit to the position. Again, with each time step the agent can take a step from the move set. In the 3-way corridor the agent is allowed to move up or down at every step or stay in its position. For a move set like in (\ref{eq:stepset}) every joint has three available steps. Therefore, the trajectory of a 1 dof joint can be described by a 3-way corridor (see also Figure \ref{fig:grid_corridor}). In \cite{ault2019counting}, the authors compute the number of paths in a corridor with Fourier analysis. The result for the number of paths with $m$ steps, starting from position $q^a$ and ending at position $q^b$, is
\vspace{-8pt}
\begin{equation}
|L^m_1|(q^a, q^b) = \frac{2}{d}\sum_{\omega = 1}^{d-1}\sin \left( \frac{\pi \omega}{d}\frac{q^b}{\delta q}\right) \left[ 1 + 2\cos \left( \frac{\pi \omega}{d}\right)\right]^m \sin \left( \frac{\pi \omega}{d}\frac{q^a}{\, \delta q}\right)
\vspace{-8pt}
\end{equation}
Positional joint limitations can be expressed with the corridor height $h=d-1 = (q^{max}-q^{min})/ \delta q$.

This solution generalizes to $n$ dimensions \cite{ault2019counting} as:
\begin{equation}
 |L^m_n|(\pmb{q}^a, \pmb{q}^b) = 
 \begin{cases}
 \!
 \begin{alignedat}{2}
  & \frac{(-1)^{\frac{n}{2}}}{\prod^n_{j=1} d_j} \sum_{\pmb{\omega} = -\pmb{d}+1}^{\pmb{d}} \cos \left( \frac{\pi \pmb{\omega}}{\pmb{d}} \frac{\pmb{q}^b}{\delta \pmb{q}}\right) \left[ \hat{T}(\pmb{\omega})\right]^m \prod_{j=1}^n \sin \left( \frac{\pi \omega_j}{d_j} \frac{q^a_j}{\delta q_j} \right) \, , \text{$n$ is even}
  \\
  & \frac{(-1)^{\frac{n+1}{2}}}{\prod^n_{j=1} d_j} \sum_{\pmb{\omega} = -\pmb{d}+1}^{\pmb{d}} \sin \left( \frac{\pi \pmb{\omega}}{\pmb{d}} \frac{\pmb{q}^b}{\delta \pmb{q}}\right) \left[ \hat{T}(\pmb{\omega})\right]^m \prod_{j=1}^n \sin \left( \frac{\pi \omega_j}{d_j} \frac{q^a_j}{\delta q_j} \right) \, , \text{$n$ is odd}
 \end{alignedat}
 \end{cases}
 \label{eq:corridor_ndim}
\end{equation}
where $\pmb{d}$ has been extended to $n$ dimensions and
\vspace{-4pt}
\begin{equation}
 \hat{T}(\pmb{\omega}) = \sum_{\pmb{\mu} \in M^+} \prod_{j=1}^{n} \left( 2 \cos \left( \frac{\pi \omega_j}{d_j}\right)\right)^{\mu_j}
 \vspace{-4pt}
\end{equation}
with $M^+ \subset M$ being the subset of moves having only nonnegative entries. The divisions between n-tuples (bold faced) are point wise divisions and their product is the usual scalar product.

\begin{example}
 Consider a 3 dof robot ($n = 3$), where each joint has a configuration range of $Q = [-135^{\circ}, 135^{\circ}]$, a resolution of $\delta q = 2^{\circ}$ and the move set $M_1 = [-\delta q, 0, \delta q]$. The robot starts at position $\pmb{q}^a = [0,0,0]$ and is supposed to reach position $\pmb{q}^b = [40^{\circ}, -20^{\circ}, 80^{\circ}]$ within $m=50$ steps. According to equation (\ref{eq:corridor_ndim}) there are $|L^{50}_3|\, ([0,0,0], [40^{\circ}, -20^{\circ}, 80^{\circ}]) = 6.15 \cdot 10^{52}$ possible paths. 
\end{example}

\vspace{-24pt}
\begin{figure}[htbp]
\centering
\includegraphics[width=0.9\textwidth]{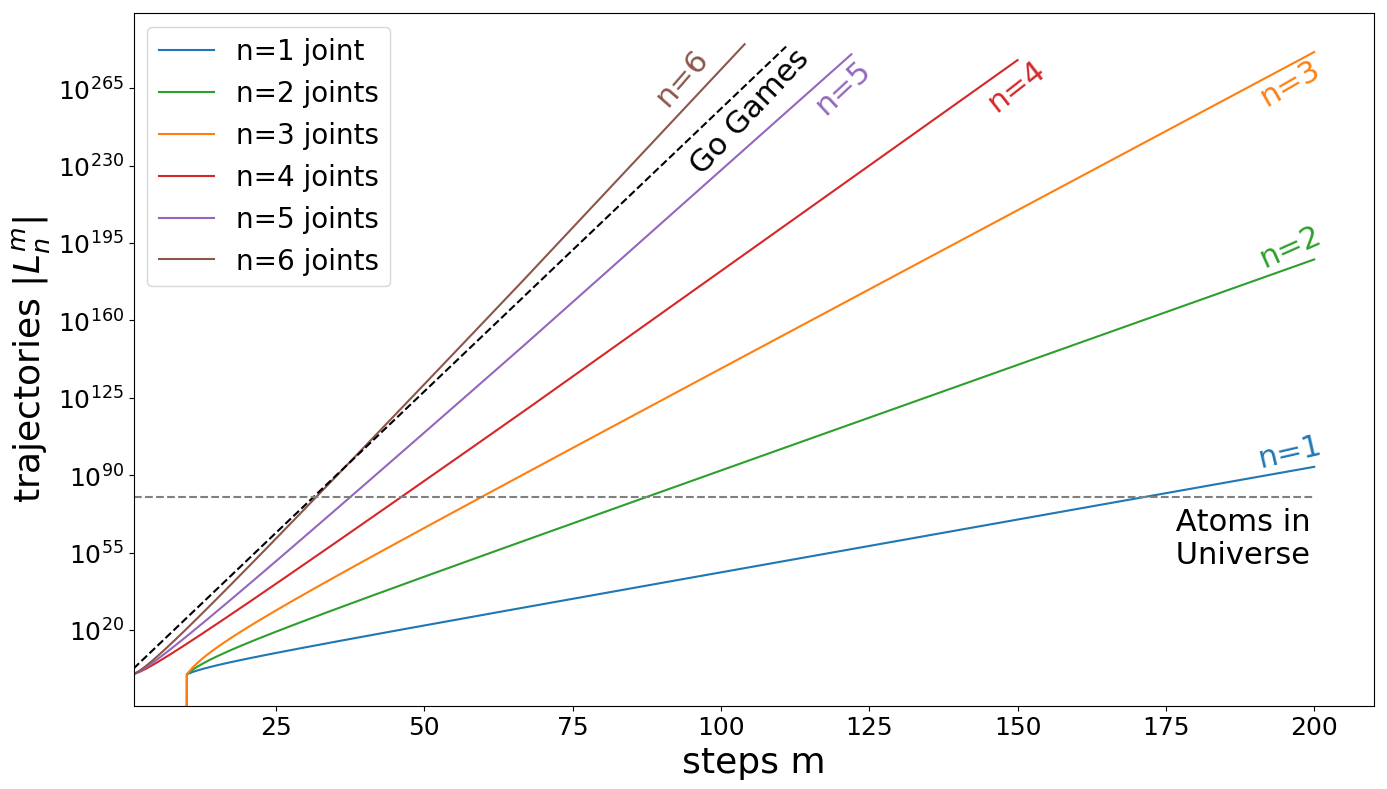}
\vspace{-6pt}
\caption{Number of paths between two configurations as a function of the number of steps for $n=1, 2, 3, 4, 5, 6$ joints (approximated for $n \geq 4$).}
\label{fig:scaling}
\end{figure}
\vspace{-18pt}
Figure \ref{fig:scaling} visualises the scaling of the number of possible trajectories against the trajectory length for $n=1, 2, 3, 4, 5, 6$ joints, which have the same specifications as in example 2. For $n \leq 3$ the number of paths has been calculated with equation (\ref{eq:corridor_ndim}). The start and end position are separated by $20^{\circ}$ in each dof (hence there are no possible paths below $m=10$ steps). For $n \geq 4$ equation (\ref{eq:corridor_ndim}) takes too long to evaluate and the paths have been estimated with $|L_n^m| = |\pmb{A}|^m / (2m+1)^n$ which is the number of different action sequences divided by the number of possible goal states. The number of atoms in the universe and an upper limit on the number of Go games with $m$ steps (moves) are plotted for comparison. For the Go games we used a board size of $19\times 19$ and the simple upper limit $361^m$. For a more accurate analysis see \cite{tromp2007combinatorics}.

In general, when taking velocity into account, a joint has more than just three options to make a step to another position. The number of available steps will be the number of atomic actions. So the number $|L^m_n|(\pmb{q}^a,\pmb{q}^b)$ is to be understood as a lower boundary on the actual number of paths from $\pmb{q}^a$ to $\pmb{q}^b$.

\section{Conclusion}

By applying methods from lattice path counting to motion planning, our work gives tangible insights into the complexity of a trajectory space which is based on a discrete configuration space. The number of trajectories from one configuration to another can precisely be calculated with equation (\ref{eq:corridor_ndim}) and scales as expected exponentially with the number of dof and steps. For $n\geq6$ the number of trajectories scales faster than the number of possible Go games. 

There are several methods to deal with this complexity. A very common tool are controllers which map the abstract motor signals to quantities that are more useful for planning such as position, velocity or total torque on a joint. Using a controller allows to work on a space which is usually more closely related to the space where the task is formulated.
Another possible approach is to use a generic capability representation instead of manually specifying a sequence of actions.  The parameters of such a reperesentation define the resulting capability. Possible tools are motion primitives such as polynomials (for example in \cite{kumar2016model}), DMP's \cite{schaal2006dynamic} or Gaussian kernel functions as used in \cite{langosz2018}.

We also contribute here by providing the building blocks of a theoretically complete set of all possible trajectories. This complete set can serve as a benchmark for these parameteric motion primitive based approaches. Going forward, we are specifically investigating whether parameter constraints can be derived, such that parametric approaches respect the fundamentally discrete nature of the \textit{c-space}.

\vspace{-8pt}
\section*{Acknowledgement}
The work presented in this paper was done in the Q-Rock project funded by the Federal Ministry of Education and Research Germany (FKZ 01IW18003).
\vspace{-8pt}
%
%
%
\addcontentsline{toc}{chapter}{Bibliography}
\bibliographystyle{splncs04}
\bibliography{references}

\end{document}